\documentclass[lettersize,journal]{IEEEtran}
\usepackage{amsmath,amsfonts}
\usepackage{amsthm}
\usepackage{algorithm}
\usepackage{algorithmic}
\usepackage{multirow}
\usepackage{multicol}
\usepackage{tabularx}
\usepackage{enumitem}
\usepackage{array}
\usepackage[caption=false,font=normalsize,labelfont=rm,textfont=rm]{subfig}
\usepackage{textcomp}
\usepackage{stfloats}
\usepackage{url}
\usepackage{verbatim}
\usepackage{booktabs}
\usepackage{graphicx}
\usepackage{cite}
\usepackage{cuted}      % 提供 strip 环境
\usepackage{caption}    % 提供 \captionof 命令
\usepackage[font=small]{caption}
\usepackage{hyperref}
\usepackage{color}
\usepackage[dvipsnames]{xcolor}
\usepackage{xcolor,colortbl}
\definecolor{tabGray}{gray}{0.94}

\theoremstyle{remark}

\usepackage{makecell}   % 用于 \makecell 命令（单元格内换行）
\usepackage{xspace}
\usepackage{threeparttable}

\begin{document}

\title{X-Distill: Cross-Architecture Vision Distillation\\for Visuomotor Learning}

\author{\fontsize{12}{16}\selectfont % 设置字体大小和行距
Maanping Shao$^{\#}$,~Feihong Zhang$^{\#}$,~Gu Zhang,~Baiye Cheng,~Zhengrong Xue,~Huazhe Xu† \vspace{1mm}\\
$^{\#}$Equal contribution\quad†Corresponding author\quad
\href{https://x-distill.github.io}{\textbf{X-Distill.github.io}\xspace}\vspace{-0.3in}% <-this % stops a space
\normalsize % 恢复默认字号
% <-this % stops a space
\thanks{This work was supported by Institute for Interdisciplinary Information Sciences, Tsinghua University, Shanghai Qi Zhi Institute and Shanghai Artificial Intelligence Laboratory.(\textit{Maanping Shao and Feihong Zhang are co-first authors, contributed equally to this work.})(\textit{Corresponding author: Huazhe Xu.})}% <-this % stops a space
\thanks{Maanping Shao, Feihong Zhang, Gu Zhang and Zhengrong Xue are with Tsinghua University, Beijing 100084, China.(e\-mail:smap24@mails.tsinghua.edu.cn, zfh24@mails.tsinghua.edu.cn, zg24@mails.tsinghua.edu.cn, xzr23@mails.tsinghua.edu.cn)}
\thanks{Baiye Cheng is with Huazhong University of Science and Technology, Wuhan 430074, China.(e\-mail:U202212594@hust.edu.cn)}
\thanks{Huazhe Xu is with the Institute for Interdisciplinary Information Sciences, Tsinghua University, Beijing 100084, China, and also with Shanghai Qi Zhi Institute, Shanghai 200030, China, as well as with Shanghai Artificial Intelligence Laboratory, Shanghai 200032, China.(e\-mail: huazhe\_xu@mails.tsinghua.edu.cn)}
}

% The paper headers
\markboth{Journal of \LaTeX\ Class Files,~Vol.~14, No.~8, August~2021}%
{Shell \MakeLowercase{\textit{et al.}}: A Sample Article Using IEEEtran.cls for IEEE Journals}

%\IEEEpubid{0000--0000/00\$00.00~\copyright~2021 %IEEE}
% Remember, if you use this you must call \IEEEpubidadjcol in the second
% column for its text to clear the IEEEpubid mark.

\maketitle

\begin{strip} 
\begin{minipage}{\textwidth}
\centering
\vspace{-100pt} % 适当调整垂直间距
\includegraphics[width=\textwidth, trim=1.6cm 12cm 8.6cm 20cm, clip]{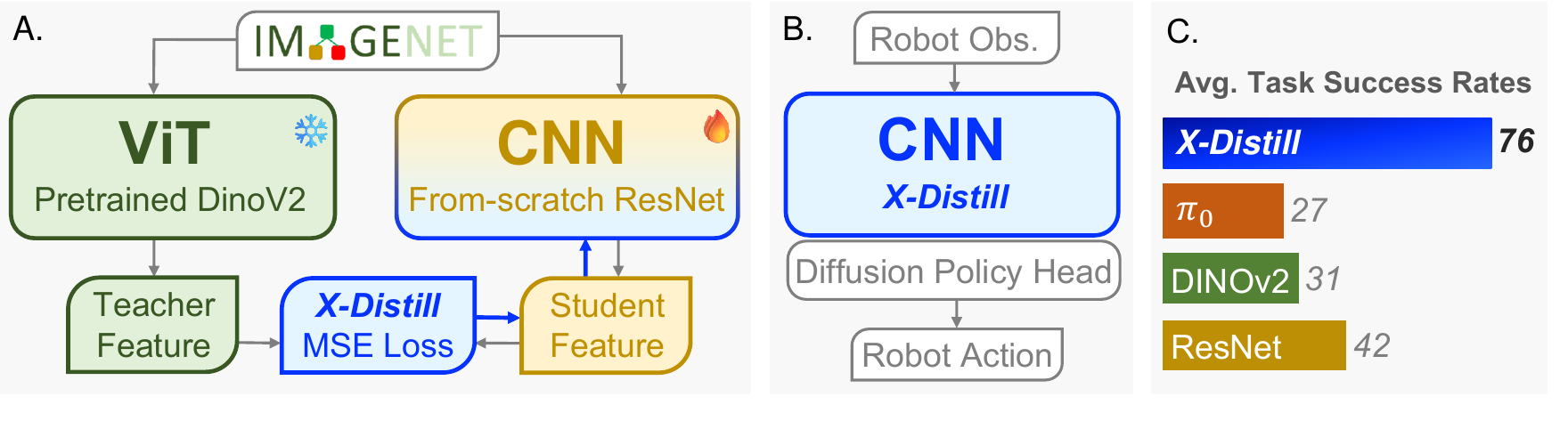}
\captionof{figure}{\textbf{X-Distill} is a simple yet effective visual encoder enabling data-efficient visuomotor learning. \textbf{A.} X-Distill is obtained by cross-architecture knowledge distillation from a large ViT teacher into a compact CNN student on general-purpose image datasets. \textbf{B.} Designed for visuomotor policy learning, X-Distill can be jointly fine-tuned end-to-end with a diffusion policy head on robotics-specific datasets. \textbf{C.} Given a few ($20\sim25$) demonstrations per task, X-Distill significantly outperforms representative counterparts on real-world manipulation tasks, exhibiting its surprising effectiveness.}
\vspace{-3mm}
\label{fig:teaser}
\end{minipage}
\end{strip}  

\begin{abstract}
Visuomotor policies often leverage large pre-trained Vision Transformers (ViTs) for their powerful generalization capabilities. However, their significant data requirements present a major challenge in the data-scarce context of most robotic learning settings, where compact CNNs with strong inductive biases can be more easily optimized. To address this trade-off, we introduce X-Distill, a simple yet highly effective method that synergizes the strengths of both architectures. Our approach involves an offline, cross-architecture knowledge distillation, transferring the rich visual representations of a large, frozen DINOv2 teacher to a compact ResNet-18 student on the general-purpose ImageNet dataset. This distilled encoder, now endowed with powerful visual priors, is then jointly fine-tuned with a diffusion policy head on the target manipulation tasks. Extensive experiments on $34$ simulated benchmarks and $5$ challenging real-world tasks demonstrate that our method consistently outperforms policies equipped with from-scratch ResNet or fine-tuned DINOv2 encoders. Notably, X-Distill also surpasses 3D encoders that utilize privileged point cloud observations or much larger Vision-Language Models. Our work highlights the efficacy of a simple, well-founded distillation strategy for achieving state-of-the-art performance in data-efficient robotic manipulation.
\end{abstract}

\begin{IEEEkeywords}
Visuomotor Policy, Knowledge Distillation, Representation Learning, Manipulation
\end{IEEEkeywords}

\section{Introduction}
\label{sec.intro}
% \IEEEPARstart{L}{egged} robots have garnered increasing attention in recent years \cite{bledt2018cheetah, bloesch2013state, bloesch2013state2}. Their leg actuators provide greater degrees of freedom compared to the wheels of traditional robots \cite{fink2020proprioceptive}, enabling them to perform more complex tasks, such as navigating rough terrains, overcoming obstacles, and operating in unstructured environments \cite{bledt2018cheetah}. Achieving these capabilities relies on accurately estimating the robot’s state, including its pose and velocity \cite{bloesch2013state}. As a result, various algorithms have been developed \cite{bloesch2013state,bloesch2013state2,fink2020proprioceptive,yang2023multi,yoon2023invariant, teng2021legged} to achieve accurate state estimation of legged robots from noisy measurements.

\IEEEPARstart{V}{isuomotor} policies, exemplified by Diffusion Policy~\cite{chi2023diffusion}, are promising solutions for generalizable robotic manipulation. As end-to-end approaches, they typically rely on a visual encoder to extract manipulation-centric features from the raw pixels of a scene, followed by a policy head that generates concrete robot actions conditioning on the extracted visual features. 

Benefiting from the success of large-scale vision pre-training~\cite{caron2021emerging,radford2021learning}, it has become a common practice in recent advances~\cite{lin2024data, xue2025demogen} to initialize the visual encoder in a visuomotor policy with off-the-shelf, pre-trained Vision Transformers (ViTs)~\cite{dosovitskiy2020vit}, e.g., CLIP~\cite{radford2021learning} or DINOv2~\cite{oquab2023dinov2}. These ViT-backend pre-trained models are found to exhibit enhanced generalization capabilities compared to Convolutional Neural Network (CNN) counterparts lacking open-world semantic knowledge, e.g., a ResNet~\cite{he2016deep} trained from scratch.

However, lacking the strong intrinsic inductive biases inherent to CNNs, such as locality and translation equivariance, ViTs are known to struggle when faced with limited amounts of training data~\cite{dosovitskiy2020vit, touvron2021training}. This issue becomes prominent and inevitable in the context of robot learning, where the dataset size is significantly smaller than in computer vision. Despite the recent trend among embodied AI startups to train policies with hundreds of \textit{hours} of high-quality data produced by a data collection factory~\cite{black2024pi_0,bu2025agibot,jiang2025galaxea}, most researchers in academia typically collect data by hand, thus favoring data-efficient policies that perform well under a dataset size constraint of tens to a few hundred manipulation \textit{trajectories}.

In this work, we find that simple advances to the visual encoder can yield higher-performing and more data-efficient visuomotor policies. More specifically, we design a \underline{cross}-architecture vision \underline{distill}ation mechanism, or \textbf{X-Distill} in short, which attempts to combine the merits of both the open-world semantic generalization capabilities of pre-trained ViT models, and the inductive bias of CNN architectures that facilitate policy optimization under the low-data regime.

On an implementation level, we instantiate X-Distill by selecting DINOv2 (ViT-L/14) as the teacher encoder, a lightweight from-scratch ResNet-18 as the student encoder, and the mean squared error (MSE) between the teacher and student features as the knowledge distillation~\cite{hinton2015distilling} loss. To make the X-Distilled encoder generally effective for diverse tasks, environments, and robot platforms, we choose the general-purpose ImageNet dataset as the distillation corpus, avoiding potential overfitting to any specific robotic scenarios.
After X-Distillation, the CNN-backend encoder with ViT pre-training knowledge can be seamlessly integrated into the policy learning pipeline, jointly fine-tuned with the policy head in an end-to-end manner on any robotics-specific datasets.

We validate the effectiveness of X-Distill by conducting experiments on $34$ simulated tasks across MetaWorld~\cite{yu2020metaworld}, Adroit~\cite{Kumar2016adroit,rajeswaran2017learningadroit}, and DexArt~\cite{bao2023dexart} benchmarks, with $10$ demonstrations per task. We also design $5$ real-world tasks, carefully defining their In-Distribution (ID) and Out-of-Distribution (OOD) conditions and preparing $20\sim25$ demonstrations per task.
Empirically, we find that Diffusion Policy with X-Distill consistently outperforms counterparts equipped with ResNet from scratch or DINOv2 as the encoder. Additionally, our policy also outperforms 3D Diffusion Policy~\cite{ze20243d}, which utilizes privileged 3D observation, as well as $\pi_0$~\cite{black2024pi_0}, a Vision-Language-Action (VLA) model that adopts a much larger VLM~\cite{beyer2024paligemma} as the visual perception encoder.
Finally, we present a detailed qualitative analysis of the learned representations, providing insights into how X-Distill achieves superior performance over the baseline methods.
\textbf{\textit{Please refer to the \href{https://x-distill.github.io}{project website} for robot videos.}}

\section{Related Works}
\label{sec:related}

\subsection{Visual Representation Learning}

\label{subsec:vis-rep-learning}

After the dominance of Convolutional Neural Networks (CNNs)~\cite{he2016deep,simonyan2014very} in the 2010s, Vision Transformers (ViTs)~\cite{dosovitskiy2020vit,touvron2021training} have gained increasing popularity in the 2020s because of their superior scaling capabilities and impressive representational power when pre-trained on large-scale datasets~\cite{caron2021emerging,he2020momentum}.
Despite this trend, CNNs maintain a crucial edge in low-data regimes and continue to see widespread practical deployment.
The key reason is their strong \textbf{inductive bias} --- the convolutional operator imposes assumptions of locality and spatial weight sharing, which make them remarkably data-efficient. 
On the other hand, ViTs lack such biases and therefore require exposure to massive datasets to learn fundamental visual concepts. 
This discrepancy in data requirements keeps CNNs popular in many specialized domains, such as medical diagnostics~\cite{shamshad2023transformers} or manufacturing quality control~\cite{liu2024deep}, where large labeled datasets are often unavailable.

% Recent works also explore adapting frozen models via task-specific adapters~\cite{sharma2023lossless, liu2023tail}. Due to the lack of public codebases, we evaluate this paradigm using LoRA~\cite{hu2022lora}. Our results (Sec.~\ref{app:r3m_lora}) show that in data-scarce regimes, our cross-architecture distillation significantly outperforms such parameter-efficient fine-tuning (PEFT) methods.

% While parameter-efficient fine-tuning (PEFT) using adapters has been proposed~\cite{sharma2023lossless, liu2023tail}, our experiments with LoRA~\cite{hu2022lora} reveal that in extremely data-scarce regimes, cross-architecture distillation yields superior sample efficiency compared to adapting large ViTs.

\subsection{Cross-Architecture Knowledge Distillation}

Knowledge distillation (KD) has become a cornerstone technique for model compression and knowledge transfer. Most work has focused on distillation between \textbf{homologous architectures}, including traditional CNN-to-CNN approaches~\cite{hinton2015distilling} and more modern ViT-to-ViT frameworks designed for efficiency such as TinyViT~\cite{wu2022tinyvitfastpretrainingdistillation}. For robotics, homologous knowledge distillation was recently explored by Theia~\cite{shang2024theia}, which fuses knowledge from multiple pre-trained ViTs into a single unified ViT encoder.
In comparison, \textbf{cross-architecture distillation} is comparatively underexplored.~\cite{liu2022crossarchitectureknowledgedistillation}. 
A representative work is DeiT~\cite{touvron2021training}, a CNN-to-ViT distillation where a CNN teacher can stabilize a data-hungry ViT student.
% , and the converse direction of ViT-to-CNN transfer which aims to imbue a computationally efficient CNN with the powerful global understanding of a large Transformer~\cite{liu2022crossarchitectureknowledgedistillation}. 
By contrast, our work adopts the converse approach: a ViT-to-CNN distillation aiming to combine the inductive bias of a CNN with the powerful semantic understanding of a large-scale pre-trained ViT.

\subsection{Visuomotor Policy Learning}

Visuomotor policy learning is a promising paradigm for robotic manipulation. 
Representative works in this vein include Diffusion Policy~\cite{chi2023diffusion} and related approaches~\cite{pearce2023imitating_diffusion, reuss2023goal, xian2023chaineddiffuser}, which typically consists of a visual encoder followed by a policy head network. 
Recently, Vision-Language-Action (VLA) models have been proposed to replace the visual encoder with more capable vision-language models (VLMs)~\cite{openvla,black2024pi_0}, enabling impressive generalization abilities such as zero-shot skill deployment in unseen homes~\cite{intelligence2025pi05visionlanguageactionmodelopenworld}. 
However, finetuning the VLM requires a substantial amount of training data. State-of-the-art VLAs such as $\pi_0$~\cite{black2024pi_0}, AgiBot GO-1~\cite{bu2025agibot}, and Galaxea G0~\cite{jiang2025galaxea} all rely on their embodiment-specific large-scale datasets, measured either in millions by the number of trajectories or in hundreds of hours by the physical on-robot execution time.
% massive datasets for pre-training. Furthermore, even after such extensive pre-training, their adaptation to specific tasks via SFT remains highly data-intensive. For example, $\pi_0$ reports that successful fine-tuning necessitates a substantial amount of task-specific data, with the simplest tasks requiring at least $5$ hours and more complex tasks demanding $100$ or more hours. This indicates that direct supervised fine-tuning of VLAs yields suboptimal performance in low-data regimes. 
In this work, we focus on training capable visuomotor policies when a limited  amount of training data is available, i.e., using only $\sim25$ demonstration trajectories per task. 
% building upon the Diffusion Policy architecture, we demonstrate that our method achieves superior performance compared to directly fine-tuned VLAs, particularly when extensive pre-training corpora and large-scale task-specific data for SFT are unavailable.

\section{Method}
\label{sec:method}
This section details the X-Distill framework, whose overall procedure is summarized in Algorithm~\ref{alg:policy_training_sub}.

\begin{algorithm}
\caption{How to acquire and leverage X-Distill.}
\label{alg:policy_training_sub}
\begin{minipage}[t]{0.48\textwidth}
    \begin{algorithmic}[1]
    \vspace{0.2em}
    \STATE \textbf{Step 1: Knowledge Distillation}
    \STATE \textbf{Input:} Teacher encoder $\mathcal{T}$ (frozen DINOv2), Student encoder $\mathcal{S}$ (from-scratch ResNet), Domain-agnostic dataset $\mathcal{D}_{\text{large}}$ (ImageNet).
    \FOR{each training epoch}
        \FOR{each batch $x \in \mathcal{D}_{\text{large}}$}
            \STATE $z_T \leftarrow f_{\mathcal{T}}(x)$.
            \STATE $z_S \leftarrow f_{\mathcal{S}}(x)$.
            \STATE $L \leftarrow \mathcal{L}_{\text{KD}}(z_S, \text{sg}(z_T))$ \ \# Eq.~\eqref{eq:kd_loss}
            \STATE Update student encoder $\mathcal{S}$ via $\nabla_{\mathcal{S}} L$.
        \ENDFOR
    \ENDFOR
    \STATE Save the weights of S as S*
    \STATE \textbf{Output:} X-Distilled encoder weights $\mathcal{S}^*$.
    \end{algorithmic}
\end{minipage}
\hfill
\begin{minipage}[t]{0.48\textwidth}
    \begin{algorithmic}[1]
    \vspace{0.4em}
    \STATE \textbf{Step 2: Policy Finetuning}
    \STATE \textbf{Input:} X-Distilled encoder weights $\mathcal{S}^*$, Diffusion policy head $\pi_{\theta}$, Domain-specific dataset $\mathcal{D}_{\text{robotics}}$.
    \STATE Initialize encoder $\mathcal{S}$ with weights from $\mathcal{S}^*$.
    \FOR{each training epoch}
        \FOR{each batch $(o, a) \in \mathcal{D}_{\text{robotics}}$}
            \STATE $z_{img} \leftarrow f_{\mathcal{S}}(x)$.
            \STATE $c \leftarrow \text{concat}(z_{img}, s)$.
            \STATE Compute $L_{\text{diff}}$. \ \# Eq.~\eqref{eq:diff_loss}
            \STATE Update $\mathcal{S}$ and $\pi_{\theta}$ via $\nabla_{\mathcal{S}, \theta} L_{\text{diff}}$.
        \ENDFOR
    \ENDFOR
    \STATE \textbf{Output:} Trained encoder and policy $(\mathcal{S}^{**}, \pi_{\theta}^*)$.
    \end{algorithmic}
\end{minipage}
\end{algorithm}

\subsection{X-Distill: A Cross-Architecture Distillation Method}
\label{sec:x-distill}
We employ cross-architecture knowledge distillation to transfer the representational capabilities of a large Vision Transformer (ViT) into a compact CNN with beneficial inductive biases.
Crucially, this entire process is conducted exclusively on the general-purpose ImageNet-1K~\cite{deng2009imagenet} dataset ($\mathcal{X}$), which contains approximately $1.3$ million images depicting a wide variety of real-world objects and scenes. This decoupling of visual feature distillation from the downstream domain-specific datasets makes X-Distill entirely \textbf{domain-agnostic}. In other words, the resulting X-Distill encoder is universally suitable for all kinds of robotic manipulation tasks, thus avoiding potential overfitting to any specific environments, camera setups, or robotic embodiments.

\paragraph{Selection of teacher and student networks.}
We select the pre-trained DINOv2 (ViT-L/14) model as our teacher $\mathcal{T}$. With approximately $304\mathrm{M}$ parameters, this large-scale model is used off-the-shelf as a frozen feature extractor, serving as a robust source of semantic and structural visual knowledge. For the student model $\mathcal{S}$, we choose a highly compact ResNet-18 architecture with only $11\mathrm{M}$ parameters. The choice of student network prioritizes not only its computational efficiency with a network parameter size nearly $28\times$ smaller than the teacher, but also its strong inductive biases such as spatial locality that are beneficial for manipulation tasks.

\paragraph{Domain-agnostic distillation.}
The student is trained to replicate the feature outputs of the teacher on ImageNet-1K. For a given input image $x$, we extract the global \texttt{[CLS]} token from the DINOv2 teacher, which serves as the target feature vector. The ResNet-18 student architecture is modified with a final linear layer to match the feature dimension of the teacher. The core objective is then to minimize the direct Mean Squared Error (MSE) between these two feature vectors:
\begin{equation}
\mathcal{L}_{\text{KD}} = \mathbb{E}_{x \sim \mathcal{X}} \left[ \left\| f_{\mathcal{T}}(x) - f_{\mathcal{S}}(x) \right\|^2_2 \right]
\label{eq:kd_loss}
\end{equation}
where $f_{\mathcal{T}}$ and $f_{\mathcal{S}}$ represent the complete feature extraction processes of the teacher and the dimension-aligned student, respectively. This process results in a ResNet-18 with parameter weights $\mathcal{S}^*$, which encodes the open-world generalization knowledge of the teacher network.

\subsection{Finetuning X-Distill for Visuomotor Policy Learning}
\label{sec:policy_learning}
Given the powerful initialization provided by X-Distill, we deploy the encoder $\mathcal{S}^*$ for downstream policy learning on a target robotics dataset. We use a Diffusion Policy~\cite{chi2023diffusion} head, which generates action chunks conditioned on robot observations.

At each timestep, the distilled encoder $\mathcal{S}^*$ processes a history of camera images $x_{t-T_o+1:t}$ into a visual feature vector $z_{\mathrm{img}}$. This vector is concatenated with the robot's proprioceptive state $s_t$ to form a comprehensive conditioning vector, $c = \text{concat}(z_{\mathrm{img}}, s_t)$. This conditioning vector $c$ guides the entire action generation process. During inference, actions are generated by iteratively denoising a random Gaussian tensor, conditioned on this vector $c$.

Crucially, both the distilled encoder $\mathcal{S}^*$ and the diffusion policy head $\pi_{\theta}$ are jointly trained on robotics-specific datasets. This end-to-end optimization allows the powerful, general-purpose features from the distillation phase to be fine-tuned and specialized for the specific demands of the manipulation task. The entire system is optimized by minimizing the diffusion loss objective:
\begin{equation}
\mathcal{L}_{\text{diff}} = \mathbb{E}_{\mathbf{A}^0, \epsilon, k} \left[ \|\epsilon - \epsilon_\theta(\mathbf{A}^0 + \sigma_k \epsilon | c, k)\|^2 \right],
\label{eq:diff_loss}
\end{equation}
where $\mathbf{A}^0$ denotes the ground-truth actions, $\epsilon \sim \mathcal{N}(0, \mathbf{I})$, and $k$ is sampled from the diffusion steps.

\begin{figure*}[t]
    \centering
    \includegraphics[        
        width=\textwidth,
        trim=0cm 8cm 0cm 0cm, % 左 下 右 上 裁剪尺寸
        clip % 启用裁剪功能
        ]{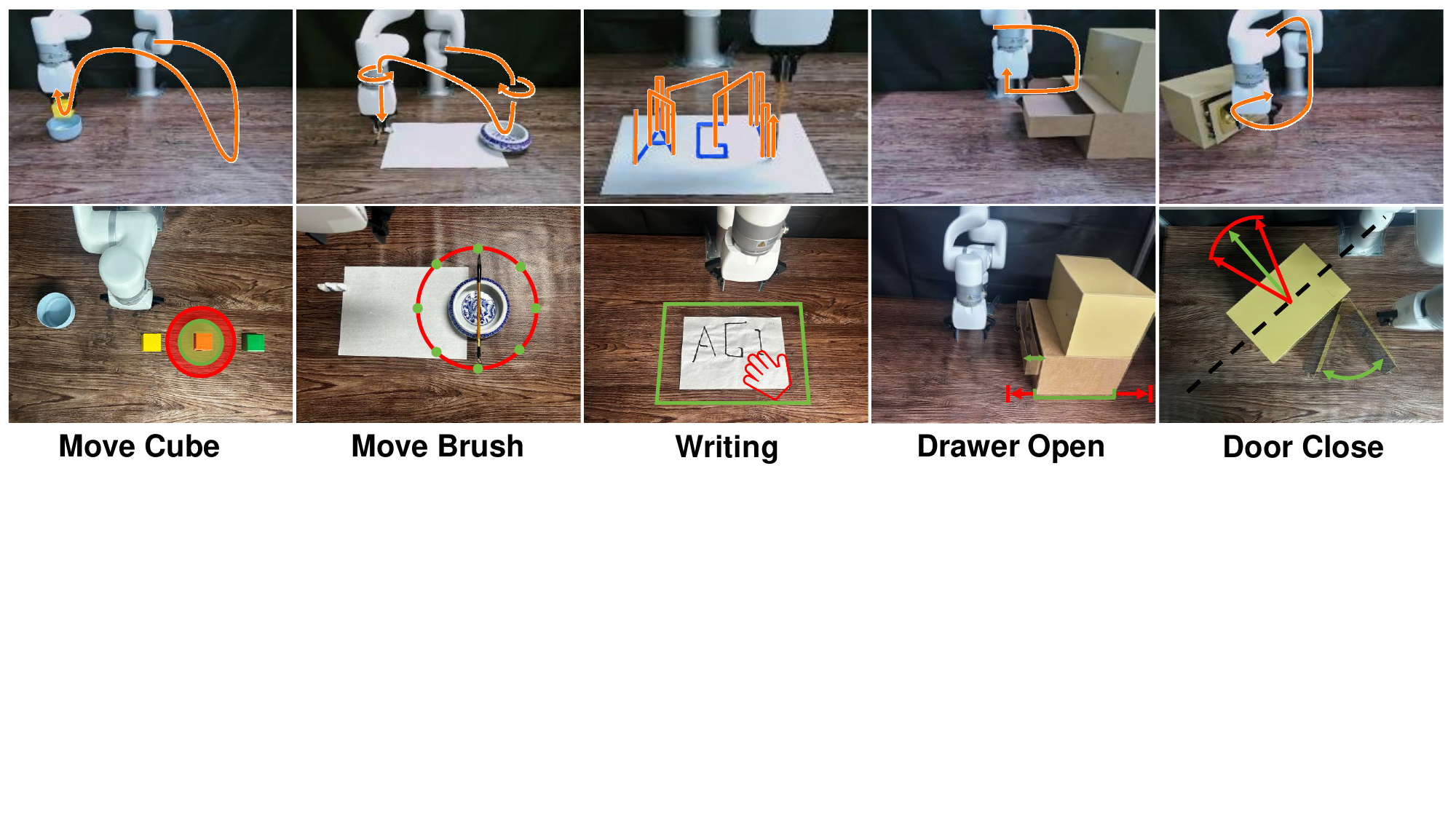}
    \caption{\textbf{Visualization of configurations for our real-world tasks.} The orange arrow provides a schematic representation of the gripper trajectory as derived from the data. The green regions represent the distribution of object/robot configurations seen during training demonstrations, while the red regions illustrate the novel configurations used for generalization testing.}
    \label{fig:ood_setup}
\end{figure*}

\begin{table*}[t]
\centering
\begin{threeparttable}
\small
\caption{\textbf{Averaged success rates on MetaWorld, Adroit and Dexart benchmarks.} PointNet-DP3 is marked in \textcolor{gray}{gray} because it processes privileged background-cropped 3D point clouds.}
\vspace{-2mm}
\label{tab:reorder}
% 调整列间距以确保表格不会过宽
\setlength{\tabcolsep}{4pt} 

% 1. 移除所有竖线 |
\begin{tabular}{l cccc c c c}
\toprule
% 2. 重构表头，使用 multirow 和 multicolumn
\multirow{2}{*}{\textbf{Method}} & \multicolumn{4}{c}{\textbf{MetaWorld}} & \multirow{2}{*}{\textbf{Adroit (3)}} & \multirow{2}{*}{\textbf{Dexart (2)}} & \multirow{2}{*}{\textbf{Average}} \\
% 3. 使用 cmidrule 创建局部横线，增强层次感
\cmidrule(lr){2-5}
& (easy 20) & (medium 7) & (hard 1) & (very hard 1) & & & \\
\midrule
ResNet-scratch             & 76.6 & \underline{48.0} & 38.0 & \underline{50.0} & 37.7 & 54.5 & 64.1 \\
DINOv2               & \underline{78.5} & 46.0 & \textbf{48.0} & 38.0 & \underline{51.7} & 58.0 & \underline{66.2} \\
Depth-Anything       & 68.2 & 29.3 & \underline{42.0} & 43.0 & 40.3 & \textbf{66.0} & 56.1 \\
Theia                & 50.9 & 13.7 & 0.0  & 38.3 & 8.7  & 24.0 & 36.0 \\
\textbf{X-Distill (Ours)} & \textbf{93.9} & \textbf{88.3} & \textbf{48.0} & \textbf{88.0} & \textbf{68.3} & \underline{63.5} & \textbf{87.2} \\
\textcolor{gray}{PointNet-DP3}               & \textcolor{gray}{90.4} & \textcolor{gray}{70.6} & \textcolor{gray}{14.0}  & \textcolor{gray}{72.0} & \textcolor{gray}{40.7}  & \textcolor{gray}{85.0} & \textcolor{gray}{84.0} \\
\bottomrule
\end{tabular}
% \begin{tablenotes}
%     \tiny % 保持字体大小一致
%     \item[1] Resnet-18 trained from scratch, sharing identical architecture with X-Distill, differing only in the weight initialization.
%     \item[2] 3D diffusion policy that processes ground-truth point clouds with a PointNet encoder, differing from other methods which only use RGB image inputs.
% \end{tablenotes}
\end{threeparttable}
\vspace{-3mm}
\end{table*}

\begin{table*}[h]
\centering
\small
\caption{\textbf{Ablation study on MetaWorld benchmarks.} We evaluate the impact of teacher model scale (DINOv2-L vs. S), student architectural bias (CNN vs. ViT), and student model scale.}
\vspace{-2mm}
\label{tab:compact}

% 可以稍微减小列间距，让表格更紧凑
\setlength{\tabcolsep}{4pt}

% lccccc 代表 1 个左对齐 + 5 个居中，共6列
\begin{tabularx}{0.82\textwidth}{c|c|ccccc} 
\toprule
% 使用 \makecell{... \\ ...} 命令来换行，让标题变窄
Teacher& Student & \makecell{MW-20 (easy)} & \makecell{MW-7 (medium)} & \makecell{MW-1 (hard)} & \makecell{MW-1 (v. hard)} & Average \\
\midrule
 & ~ResNet-18 (11M) & \underline{93.9} & \textbf{88.3} & \underline{48.0} & \underline{88.0} & \textbf{90.7} \\
DINOv2-L & ViT-S-Half (11M) & 72.0 & 25.3 & 2.0 & 40.0 & 57.2 \\
 & ~ConvNeXt (89M) & 91.8 & 77.4 & \textbf{50.0} & 83.0 & 86.6 \\
 \midrule
DINOv2-S & ~ResNet-18 (11M) & \textbf{94.3} & \underline{87.3} & 43.0 & \textbf{90.0} & \underline{90.6} \\
\bottomrule
\end{tabularx}
\vspace{-3mm}
\end{table*}

\section{Simulation Experiments}
\label{sec:method}
\subsection{Setup}

\textbf{Simulation benchmarks.} To thoroughly evaluate the effectiveness of our method, we conduct experiments across a total of $34$ tasks from $3$ distinct MuJoCo-based robotic manipulation benchmarks. Our evaluation encompasses tasks requiring parallel gripper manipulation from MetaWorld~\cite{yu2020metaworld}, dexterous motor skills from Adroit~\cite{Kumar2016adroit,rajeswaran2017learningadroit}, and articulated object manipulation from DexArt~\cite{bao2023dexart}.
% These tasks are built on different simulation frameworks, ensuring our benchmarking is not constrained by a specific simulator. 
Tasks in MetaWorld are categorized into various difficulty levels---\textit{easy}, \textit{medium}, \textit{hard}, and \textit{very hard}---based on~\cite{seo2023mwm}. 

\textbf{Expert demonstrations.} $10$ trajectories are collected for each simulation task. For MetaWorld, scripted policies are employed. Trajectories in the remaining domains are gathered using agents trained via reinforcement learning (RL): specifically, VRL3~\cite{wang2022vrl3} is applied for Adroit, while PPO~\cite{schulman2017ppo} is utilized for the remaining benchmarks.
% We generate successful trajectories with these RL agents and ensure a consistent set of demonstrations is used for all imitation learning methods.

\textbf{Evaluation Metric.} We report all results averaged over $3$ random seeds ($0$, $1$, and $2$). For each individual training run, we evaluate the policy on $20$ episodes every $200$ epochs, and the highest success rate achieved throughout the run is reported for that seed. The final values presented in our tables are the mean of these scores across the $3$ seeds.

\subsection{Performance}

\textbf{Compared methods.} We compare X-Distilled ResNet-18 ($11\mathrm{M}$) against several visual encoder counterparts with a similar number of parameters, including:
\begin{itemize}[leftmargin=*, nosep, topsep=-1pt]
    \item \textbf{ResNet-scratch}~\cite{he2016deep}, ResNet-18 ($11\mathrm{M}$) trained from scratch;
    \item \textbf{DINOv2}~\cite{oquab2023dinov2}, ViT-small ($21\mathrm{M}$) pre-trained using large-scale self-supervision;
    \item \textbf{Depth-Anything}~\cite{yang2024depth}, ViT-small ($24.8\mathrm{M}$) trained for monocular depth estimation;
    \item \textbf{Theia}~\cite{shang2024theia}, ViT-small ($22\mathrm{M}$) that distills multiple vision foundation models.
\end{itemize}

Additionally, we also benchmark against \textbf{PointNet-DP3}~\cite{ze20243d}, a PointNet-based architecture ($0.06\mathrm{M}$) processing privileged background-cropped 3D point cloud observations.

\textbf{Main results.} As summarized in Table~\ref{tab:reorder}, X-Distill achieves the best overall average performance across all $34$ tasks. It consistently outperforms all 2D vision baselines by a significant margin, securing state-of-the-art success rates in most simulation benchmarks. This validates the effectiveness of our distillation strategy for data-scarce visuomotor learning.

Notably, our 2D approach remains highly competitive even in geometrically demanding settings where methods leveraging privileged 3D inputs have a natural advantage. For instance, the DexArt-Toilet task requires the robot to lift the toilet lid from a frontal viewpoint, which is inherently challenging to estimate the depth relationship between the gripper and the object to be manipulated from a single RGB image. Nevertheless, X-Distill still demonstrates decent performance in many of these challenging tasks, showcasing a strong prior in spatial reasoning.

\begin{figure*}[t!]
    \centering
    \includegraphics[        
        width=\textwidth,
        trim=0.5cm 1.5cm 0.4cm 1.0cm, % 左 下 右 上 裁剪尺寸
        clip % 启用裁剪功能
        ]{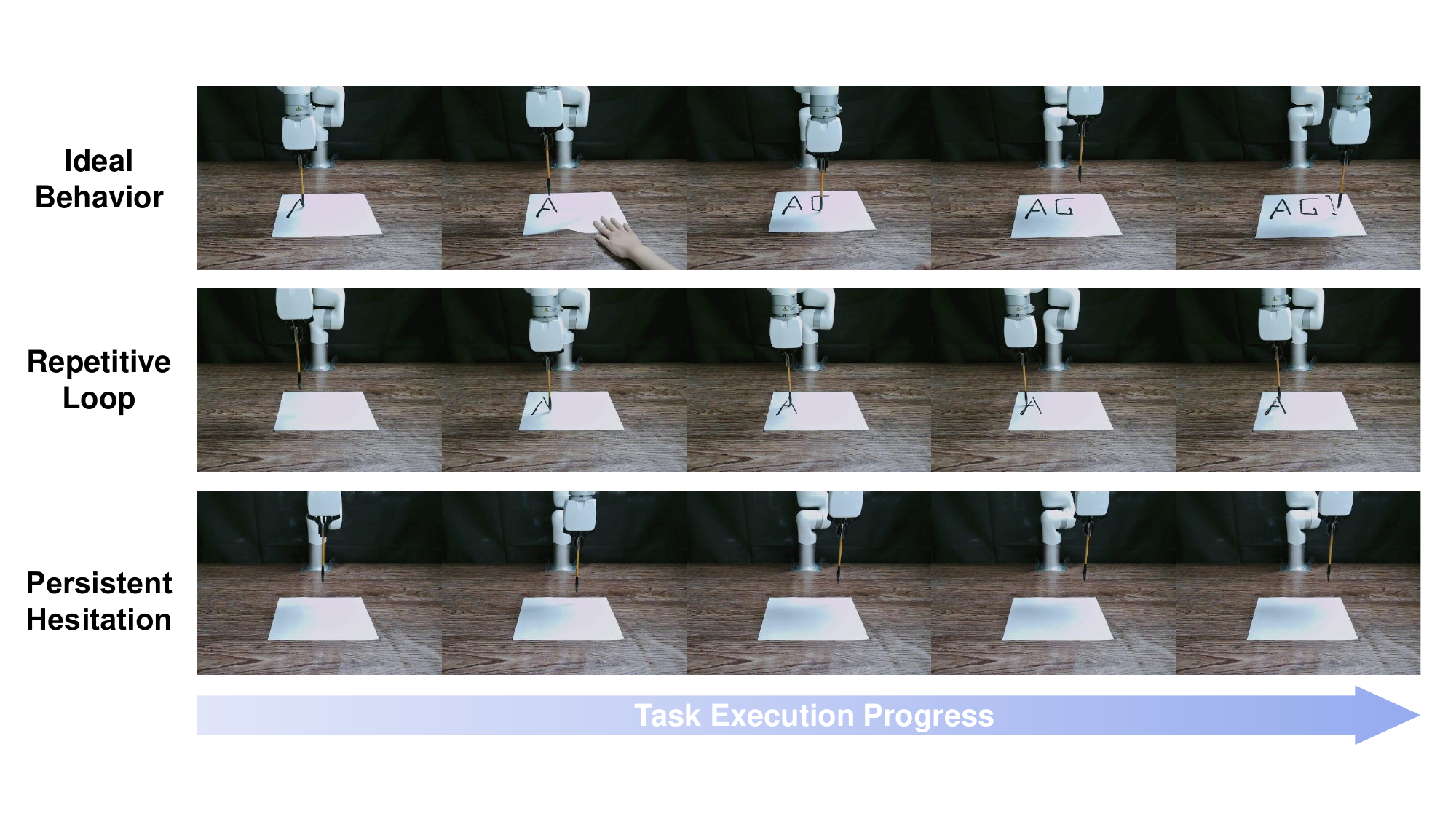}
    \caption{\textbf{Representative trajectory types observed in the ``Writing AGI'' task.} We identify three distinct behaviors: \textbf{(1) Ideal Behavior:} Successful and robust execution of all three letters, even under perturbation. \textbf{(2) Repetitive Loop:} Perseverative behavior where the policy gets stuck repeatedly writing the first letter `A'. \textbf{(3) Persistent Hesitation:} Dithering motion above the paper without initiating the writing task.}
    \label{fig:sequence}
\end{figure*}

% \textbf{Impact of Vision Architecture and Model Scale.} 

\begin{table*}[t]
\centering
\small
\caption{\textbf{A comparison of task execution success rates (\%) for real-world tasks}, along with the numbers of demonstrations and evaluation trials.}
\label{tab:real_world}
% 缩小列间距以适应页面
\setlength{\tabcolsep}{3.2pt} 

\begin{tabular}{lccccccccccc c}
\toprule
% --- Header Part 1: Task Names & Average Title ---
% Average 标题现在只与它下面的数据对齐
\multirow{2}{*}{\textbf{Method}} & \multicolumn{3}{c}{\makecell{Move Cube}} & \multicolumn{2}{c}{\makecell{Move Brush}} & \multicolumn{2}{c}{Writing ``AGI''} & \multicolumn{2}{c}{\makecell{Drawer Open}} & \multicolumn{2}{c}{\makecell{Door Close}} & \multirow{2}{*}{\textbf{Average}}  \\
\cmidrule(lr){2-12} % 这条线只画在任务列下方，不到Average列
% --- Header Part 2: Sub-Task Names ---
& ID & OOD & C-Gen & ID & OOD & ID & OOD & ID & OOD & ID & OOD & \\
\midrule
% --- Setup Info Part (with a clear separation) ---
\# Demos & 20 & 0 & 0 & 24 & 0 & 25 & 0 & 20 & 0 & 20 & 0 & -- \\
\# Eval Trials & 15 & 5 & 10 & 4 & 8 & 5 & 4 & 5 & 15 & 5 & 10 & -- \\
% \midrule[\heavyrulewidth] % 使用一道粗线，彻底将设置与结果分开
\midrule
% 所有数据行已按照新顺序重新排列
ResNet-scratch    & 66.7 &  0.0 & 50.0 &  0.0 &  0.0 & 40.0 & 25.0 & \textbf{100.0} & 13.3 & 60.0 & 80.0 & 41.9 \\
DINOv2            & 26.7 & 20.0 & 20.0 &  0.0 &  0.0 &  0.0 &  0.0 &  80.0 & 13.3 & \textbf{100.0} & 90.0 & 31.4 \\
$\pi_0$ (SFT)      &  0.0 &  0.0 &  0.0 & 25.0 &  0.0 &  0.0 &  0.0 &  80.0 & 33.3 &  80.0 & 90.0 & 26.7 \\
\textbf{X-Distill (Ours)}   & \textbf{93.3} & \textbf{40.0} & \textbf{70.0} & \textbf{75.0} & \textbf{25.0} & \textbf{100.0} & \textbf{100.0} & \textbf{100.0} & \textbf{53.3} & \textbf{100.0} & \textbf{100.0} & \textbf{75.6} \\
\bottomrule
\end{tabular}
\end{table*}

\subsection{Ablation Studies}
We conduct ablation studies to investigate the impact of the teacher network parameter size, as well as the student network architectural bias and parameter size within our X-Distill framework. The ablation results are summarized in Table~\ref{tab:compact}.

\textbf{Teacher network parameter size.} We distill \textbf{DINOv2-S} ($21\mathrm{M}$) and \textbf{DINOv2-L} ($304\mathrm{M}$) teachers into the same ResNet-18 student. No significant difference can be observed between DINOv2-S and DINOv2-L, indicating our X-Distill framework is insensitive to the specific network configurations of a well-pre-trained teacher network. Nevertheless, we use the DINOv2-L teacher for all subsequent experiments to ensure the maximized knowledge quality that the teacher could provide.

\textbf{Student network architectural bias.} We distill the same DINOv2-L teacher into a \textbf{ResNet-18 ($11\mathrm{M}$)} and a customized \textbf{ViT-S-Half ($11\mathrm{M}$)} of the same size. 
The ResNet-18 student substantially outperforms its ViT counterpart by $33.5\%$. This highlights the crucial role of convolutional inductive biases for visuomotor learning in a low-data regime, supporting our primary hypothesis.

\textbf{Student network parameter size.} We compare our compact \textbf{ResNet-18 ($11\mathrm{M}$)} student to a much larger \textbf{ConvNeXt ($89\mathrm{M}$)}~\cite{liu2022convnet2020s} CNN counterpart. Despite its greater capacity, the larger model achieves a slightly degraded success rate by $4.1\%$ on robotics tasks. This confirms our intuition that smaller visual encoders with stronger inductive biases are easier to optimize, thus beneficial for data-efficient policy learning.

\section{Real-World Experiments}
\label{sec:exp}

\subsection{Experiment Setup}

We conduct all real-world experiments with an X-Arm 6 robotic arm, capture image observations through a web camera at $15\textrm{Hz}$, and prepare a small collection of demonstrations ($20\sim25$) per task via Meta-Quest VR teleoperation.
We design $5$ tabletop manipulation tasks, and carefully define their In-Distribution (ID) and Out-of-Distribution (OOD) object randomization ranges for rigorous and repeatable evaluation. Task execution trajectories as well as ID and OOD ranges are illustrated in Figure~\ref{fig:ood_setup}.
Detailed numbers of demonstrations and evaluation trials can be found in Table~\ref{tab:real_world}.

% For rigorous evaluation, test policy generalization across five distinct manipulation tasks. For each task, we define specific In-Domain (ID) and Out-of-Domain (OOD) conditions, as illustrated in Figure~\ref{fig:ood_setup}. The detailed data collection and evaluation protocol for each task is as follows:

More specifically, \textbf{Move Cube} requires the robot to pick up an orange cube and place it into a bowl. In addition to testing on OOD cube positions, we also conduct a color generalization (C-Gen) test with unseen yellow and green cubes.
\textbf{Move Brush} requires the robot to pick up a brush pen with various initial orientational and translational offsets and place it onto a stand.
% We collect $24$ ID demonstrations, covering $8$ rotation angles and $3$ translational offsets of the brush. We conduct $4$ ID evaluation trials and $8$ OOD trials with both orientational and translational OOD offsets.
% consists of 12 trials where the brush is placed at a fixed, seen or unseen orientation (12 o'clock direction) with randomized translational offsets.
\textbf{Writing ``AGI''} requires the robot to sequentially write letters ``AGI'' on a randomly placed piece of paper. We conduct OOD dynamic perturbation trials, where human perturbators randomly drag the paper elsewhere while the robot is writing letters.
% \paragraph{Writing.} We collect 25 demonstrations of the robot writing the word 'AGI', with the paper's position varying within an ID region. Evaluation consists of 5 trials at ID positions. The OOD evaluation is a dynamic perturbation test: during 4 separate trials, we physically drag the paper while the robot is writing to assess its reactive capabilities.
\textbf{Drawer Open} requires the robot to insert its finger into varying initial gaps of the randomly placed drawer, and then open it by sliding outward. 
% We collect $20$ ID demonstrations, and conduct $5$ ID and $15$ OOD evaluation trials.
% \paragraph{Drawer Open.} We collect 20 demonstrations of opening a drawer, where the drawer's initial closed state has minor variations. We evaluate on 20 trials in total, distributed across 1 ID and 3 OOD drawer positions. For each position, the policy is tested against 5 different initial states of the drawer.
\textbf{Door Close} requires the robot to close the door from various initial open angles by pushing it inward. 
% We collect $20$ ID demonstrations, and conduct $5$ ID and $10$ OOD evaluation trials.

% \paragraph{Door Close.} Similar to the drawer task, we collect 20 demonstrations. Evaluation is performed over 15 trials, comprising 1 ID and 2 OOD door positions, with each position tested against 5 different initial opening angles.

\subsection{Experiment Results and Analysis}

\begin{figure*}[t]
\centering

\includegraphics[width=0.99\textwidth,        
        trim=3.3cm 1cm 2.8cm 2.5cm, % 左 下 右 上 裁剪尺寸
        clip]{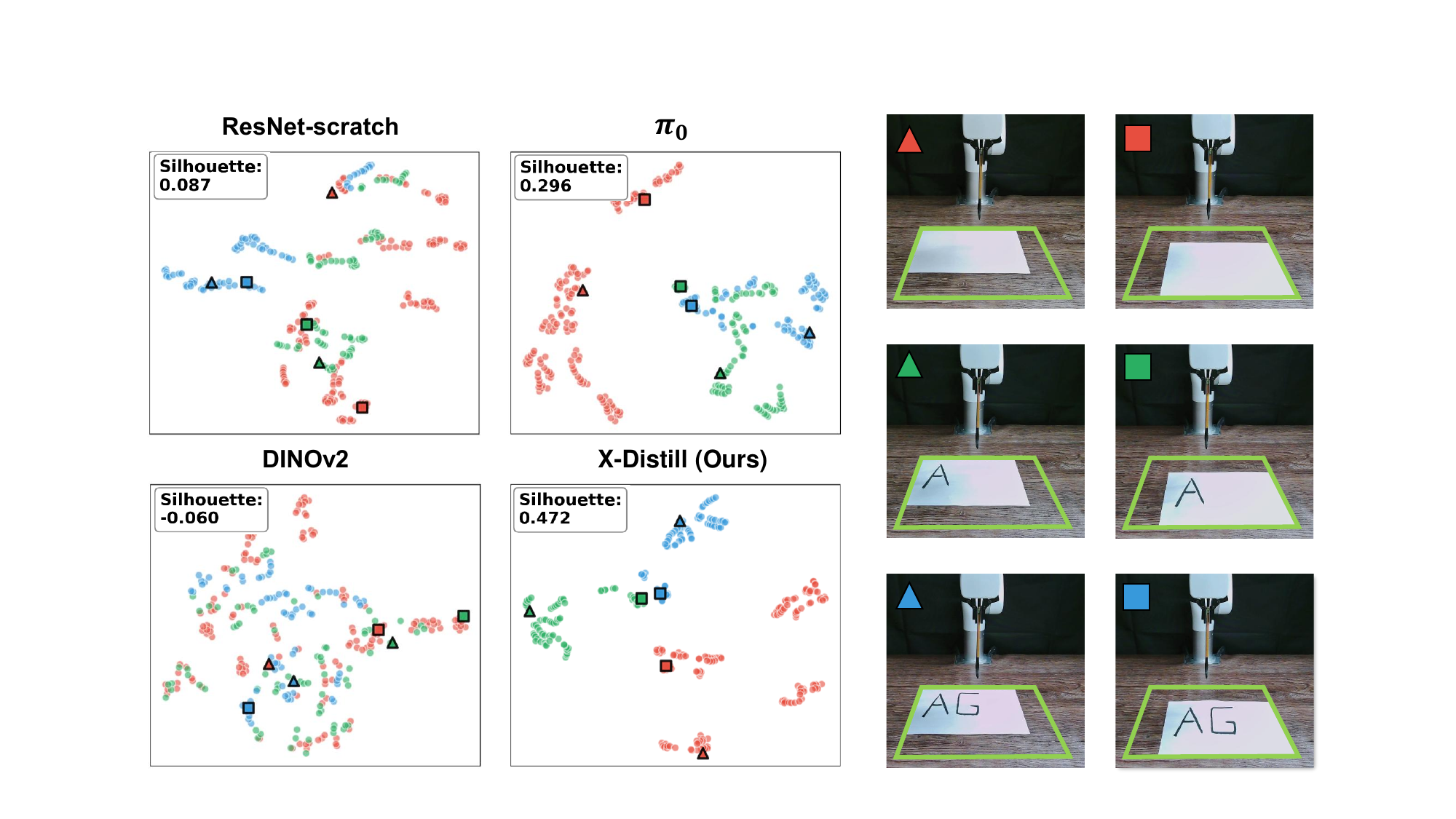}
\caption{\textbf{t-SNE visualization of learned feature spaces on the ``Writing AGI'' task.} 
Our X-Distill encoder learns to form three distinct clusters corresponding to the task's semantic stages, quantitatively confirming a well-separated feature space with a high Silhouette Score~\cite{rousseeuw1987silhouettes} of $0.472$, which indicates a high degree of cluster cohesion and separation compared with the baselines. This semantic separability is crucial for the policy to accurately identify the current task stage, enabling precise long-horizon planning for the sequential writing task.}
\label{fig:tsne_analysis}
\end{figure*}

\textbf{Baselines.} We compare our X-Distill encoder against three representative counterparts. The first two are Diffusion Policies equipped with either a ResNet encoder trained from scratch or an off-the-shelf DINOv2 encoder. Both of the two baseline policies, as well as our approach, are trained for $1500$ epochs on our task-specific data. Our third baseline is the state-of-the-art Vision-Language-Action (VLA) model, $\pi_0$~\cite{black2024pi_0}. Considering its significant computational requirements, we performed supervised fine-tuning (SFT) for $30,000$ steps, following the official recommendations, which took approximately $20$ hours on a single A100 GPU. All methods were trained using the same dataset consisting of the same limited number of demonstrations to ensure a fair comparison.

\textbf{Main results.}
The quantitative results for real-world experiments are summarized in Table~\ref{tab:real_world}. X-Distill demonstrates clear superiority, consistently outperforming all baseline approaches by a large margin and achieving the highest success rates across both ID and OOD evaluation settings. 
Simply finetuning a large ViT encoder like DINOv2 yields poor performance, confirming the challenge of effectively optimizing large Transformer networks in data-scarce scenarios and underscoring the effectiveness of our cross-architecture distillation.

The performance gap is particularly insightful when comparing against the VLA model, $\pi_0$. While $\pi_0$ shows reasonable success on simpler tasks like \textbf{Drawer Open}, it struggles significantly on more complex, high-precision tasks such as \textbf{Writing ``AGI''}, where its performance drops to zero. This suggests that directly finetuning a large, generalist VLA on small, task-specific datasets is a significant challenge. In contrast, our X-Distill framework effectively bridges this gap by transferring knowledge into a compact, data-efficient architecture, highlighting the importance of matching the model and pre-training strategy to the available data resources.

% \begin{figure*}[h]
%     \centering
%     \includegraphics[        
%         width=1.0\textwidth,
%         trim=0.5cm 1.5cm 0.4cm 1.0cm, % 左 下 右 上 裁剪尺寸
%         clip % 启用裁剪功能
%         ]{RAL/figures/sequence.pdf}
%     \caption{\textbf{Representative trajectory types observed in the ``Writing AGI'' task.} We identify three distinct behaviors: \textbf{(1) Ideal Behavior:} Successful and robust execution of all three letters, even under perturbation. \textbf{(2) Repetitive Loop:} Perseverative behavior where the policy gets stuck repeatedly writing the first letter `A'. \textbf{(3) Persistent Hesitation:} Dithering motion above the paper without initiating the writing task.}
%     \label{fig:sequence}
% \end{figure*}

\subsection{Qualitative Analysis}
\definecolor{myCustomRed}{RGB}{226, 76, 62}
\definecolor{myCustomGreen}{RGB}{42, 175, 98}
\definecolor{myCustomBlue}{RGB}{55,153,219}

We focus on our most challenging long-horizon task, \textbf{Writing ``AGI''}, where success critically depends on the encoder's ability to discern subtle but crucial visual state changes. For instance, before starting to write `G', the robot's physical state is nearly identical to when it starts writing `A'; the only distinguishing information is the visual context of the letter `A' already present on the paper. 

Empirically, we observe that Diffusion Policy with ResNet-scratch often fails by repeatedly executing the trajectory for `A', indicating an inability to visually differentiate these critical semantic states. Meanwhile, Diffusion Policy with DINOv2 and $\pi_0$ (SFT) often get stuck and start trembling before writing any letter, which is a potential sign of underfitting. In comparison, Diffusion Policy with X-Distill is the only one among the compared methods that manages to differentiate all critical stages and completes writing all the three letters sequentially one by one. Even under severe external disturbances that drag the paper away during the writing process, the X-Distill-empowered policy is still robust, responsively following the movement of the paper and rapidly adapting to the correct position for writing the next letter. Example trajectories are shown in Figure~\ref{fig:sequence}. To provide deeper insights into how X-Distill achieves superior quantitative performance compared to its counterparts, we conduct further t-SNE analysis and saliency map visualization of the learned visual representations. 

\textbf{t-SNE visualization of feature space separability.}
The global structure of the learned feature space is visualized via t-SNE~\cite{maaten2008visualizing} in Figure~\ref{fig:tsne_analysis}.
Each data point corresponds to the feature of a frame sampled from three crucial stages for policy decision making marked in three respective colors: \textcolor{myCustomRed}{(1) before writing `A'}, \textcolor{myCustomGreen}{(2) before writing `G'}, and \textcolor{myCustomBlue}{(3) before writing `I'}. The features produced from an ideal visual encoder should form three distinct clusters, corresponding to the three colors.
It can be observed that X-Distill gives a more separable feature space than the Paligemma~\cite{beyer2024paligemma} encoder extracted from $\pi_0$, while the features from both ResNet-scratch and DINOv2 are nearly indistinguishable. These results indicate that X-Distill learns a feature space that is semantically coherent and robust to visual distractors.
% Silhouette score
% The features from our X-Distill encoder form three distinct, well-separated clusters, corresponding perfectly to the task stages. This is quantitatively supported by a strong Silhouette score of 0.435. In stark contrast, the features from both the ResNet (from scratch) and DINOv2 baselines are largely indistinguishable, with near-random (0.003) or negative (-0.065) scores, respectively. 

\begin{figure*}[t!]
\centering
\includegraphics[width=0.98\linewidth,        
        trim=1cm 9cm 1cm 1cm, % 左 下 右 上 裁剪尺寸
        clip]{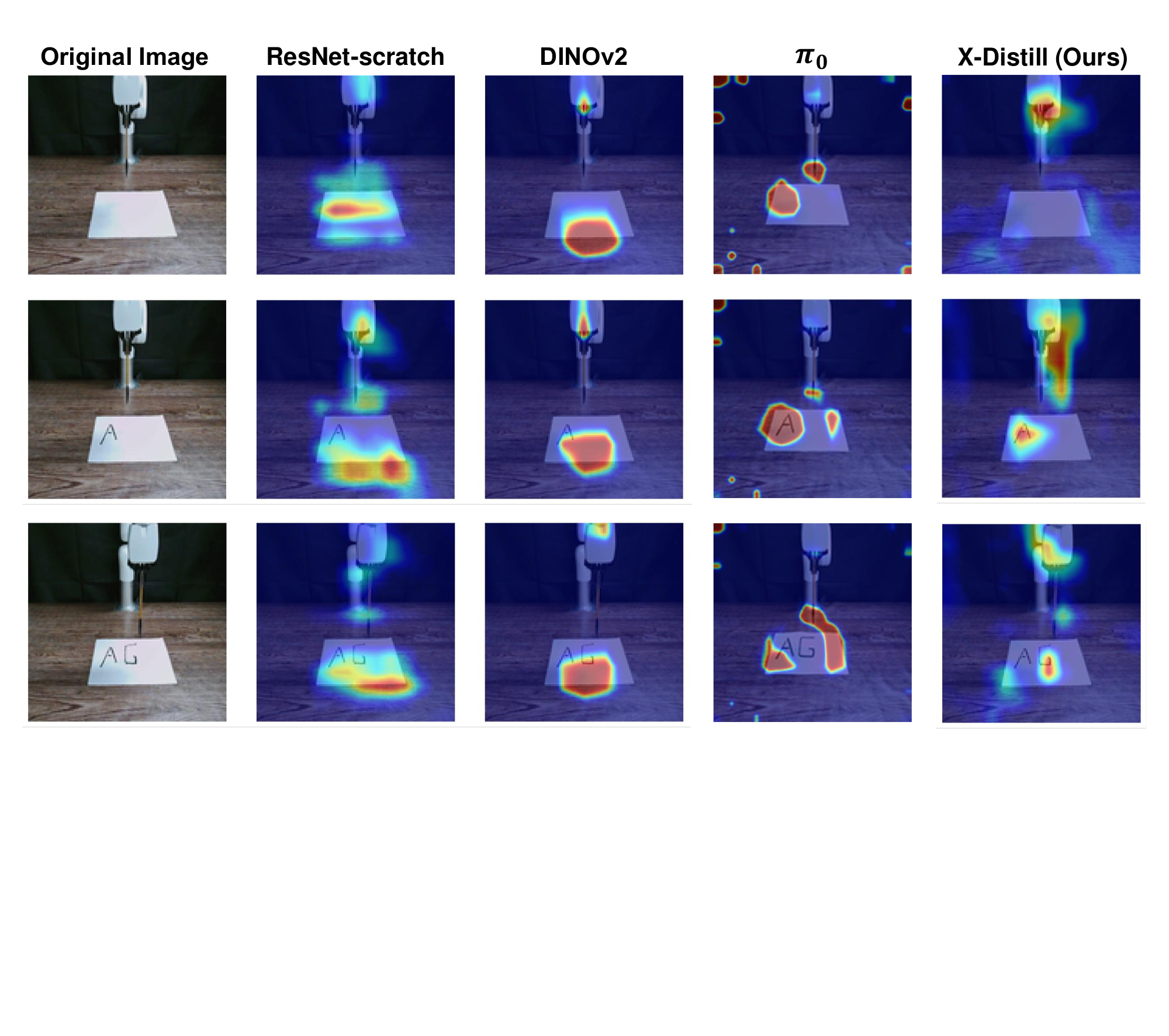}
\caption{\textbf{Saliency map comparison on the ``Writing AGI'' task.} We visualize the model's visual focus at the beginning of each writing stage. Our X-Distill encoder correctly shifts its attention from the gripper (before `A'), to the letter `A' (before `G'), and finally to the letter `G' (before `I'). Baseline models exhibit diffuse or irrelevant attention.}
\label{fig:saliency_analysis}
\end{figure*}

\textbf{Inspecting task-relevant feature attribution via saliency maps.}
To further investigate how our X-Distill achieves emergent semantic feature separation, we inspect pixel-level feature attribution by visualizing the saliency maps.
For saliency visualization, we adopt Grad-CAM for CNN-based models~\cite{Selvaraju_2019}, and the cross-attention strengths between the \texttt{[CLS]} token and all local patch features for ViT-based models~\cite{dosovitskiy2020image,chefer2021transformerinterpretabilityattentionvisualization}, providing a cross-architecture comparison of the visual focus.
As shown in Figure~\ref{fig:saliency_analysis}, both DINOv2 and $\pi_0$ are unable to effectively shift the high-attention regions throughout  the task progress, which cross-verifies our earlier judgment of underfitting. Meanwhile, the saliency maps of ResNet-scratch and X-Distill exhibits more reasonable shifting patterns, but the latter is significantly more precise. 
% In contrast, X-Distill demonstrates a remarkable ability to ground its features in task-relevant visual cues. 
More specifically, before writing `A' on the blank page, the full attention of X-Distill is focused on the \textbf{robot gripper}, the primary actor.
Then, before writing `G', its focus dynamically shifts to the \textbf{letter `A'} already on the paper.
Finally, before writing `I', X-Distill shifts attention again, attending to the \textbf{letter `G'}, whose appearance serves as the cue to write the final letter `I'.

The t-SNE and saliency map visualizations combined reveal that X-Distill successfully learns a semantically meaningful and robust visual representation. Such representation can well differentiate critical states and dynamically focus on task-relevant visual cues, which ultimately contributes to the policy's success in complex long-horizon manipulation tasks.
% which provides intuitive evidence explaining X-Distill's ability to differentiate critical states and dynamically focus on task-relevant visual cues.

% \begin{figure}[h]
% \centering
% \includegraphics[width=1.0\linewidth, trim=0cm 0cm 0cm 1cm, 
% % 左 下 右 上 裁剪尺寸 
% clip]{RAL/figures/data_scaling_plot.png}
% \caption{\textbf{Data scaling efficiency analysis.} X-Distill (Blue) demonstrates superior data efficiency, saturating at 100\% success with only 80 demos. ViT-based baselines start slowly, and generalist VLAs ($\pi_0$) show instability.}
% \label{fig:data_scaling}
% \end{figure}

% \subsection{Data Scaling Analysis}
% \label{app:data_scaling}

% To understand the scalability of our method and identify the potential crossover point where ViTs might outperform CNNs, we conducted a data scaling experiment on the real-world ``Move Brush'' task. We trained policies using $N=\{20, 40, 60, 80, 100\}$ demonstrations. The results are visualized in Figure~\ref{fig:data_scaling}.

% X-Distill demonstrates extreme data efficiency, yielding policies with $>50\%$ success rate at a cost of as few as 20 demos and reaching perfect performance at 80 demos. In contrast, the ViT baseline (DINOv2) requires significantly more data to begin learning, only reaching 41.7\% at 100 demos. This trend suggests our X-Distill is consistently competitive in low-data regime, and the crossover point where a ViT outperforms our distilled CNN might be located at hundreds or even thousands of trajectories.
% far beyond the typical data regime for accessible robotic learning.

\section{Conclusion}
\label{sec:conclusion}

We introduced \textbf{X-Distill}, a framework addressing the trade-off between ViT generalization and CNN sample efficiency. By distilling DINOv2 features into a ResNet-18 on ImageNet, we obtain a robust encoder for data-scarce robotics. Extensive experiments on $34$ simulated and $5$ real-world tasks show X-Distill outperforms standard baselines and even privileged 3D or VLA policies. Our analysis attributes this to the learned semantically separable feature space. Ultimately, X-Distill demonstrates that a simple, well-founded distillation strategy is a key enabler for data-efficient visuomotor learning. We believe X-Distill provides a practical and effective pathway towards building capable visuomotor policies with limited data, and hope it will inspire further research on cross-architecture knowledge distillation for robotics.

\textbf{Limitations and Future Work.}
While effective, our direct feature distillation leaves room for exploration. Future directions include adopting sophisticated techniques to align intermediate features~\cite{liu2022crossarchitectureknowledgedistillation} and distilling from multimodal VLA teachers to incorporate language priors. Additionally, while we focus on the data-scarce regime, investigating X-Distill's scalability in data-rich scenarios and its application to dynamic tasks like mobile manipulation remain important open questions.

\bibliographystyle{IEEEtran}
\bibliography{ref}

%\vfill

\end{document}